\documentclass[runningheads, a4paper]{llncs}
\usepackage{amssymb, amsmath, bm, latexsym, comment}
\usepackage{multirow}
\usepackage{xcolor}
\usepackage{graphicx}
\usepackage{subfigure}
\usepackage{indentfirst}
\usepackage{setspace}
\usepackage{verbatim}
\usepackage{array}
\usepackage{cite}
\usepackage{booktabs}
\usepackage{hyperref}
\usepackage[capitalize]{cleveref}
\usepackage{arydshln}
\usepackage{multirow}

\newcolumntype{L}[1]{>{\raggedright\let\newline\\\arraybackslash\hspace{0pt}}m{#1}}
\newcolumntype{C}[1]{>{\centering\let\newline\\\arraybackslash\hspace{0pt}}m{#1}}
\newcolumntype{R}[1]{>{\raggedleft\let\newline\\\arraybackslash\hspace{0pt}}m{#1}}

\begin{document}

\title{A low-rank representation for unsupervised registration of medical images}

\author{Dengqiang Jia \inst{1} \and
		Shangqi Gao \inst{2} \and
		Qunlong Chen \inst{3} \and
		Xinzhe Luo \inst{2} \and
		Xiahai Zhuang\inst{2,*}}
	
	
\authorrunning{Jia et al.}
\institute{School of Naval Architecture, Ocean and Civil Engineering, Shanghai Jiao Tong University, Shanghai, China,
	\and School of Data Science, Fudan University, Shanghai,
	\and School of Mechanical Engineering, Shanghai Jiao Tong University, Shanghai, China
	}

\maketitle

\begin{abstract}
Registration networks have shown great application potentials in medical image analysis.
However, supervised training methods have a great demand for large and high-quality labeled datasets, which is time-consuming and sometimes impractical due to data sharing issues.
Unsupervised image registration algorithms commonly employ intensity-based similarity measures as loss functions without any manual annotations.
These methods estimate the parameterized transformations between pairs of moving and fixed images through the optimization of the network parameters during training.
However, these methods become less effective when the image quality varies, e.g., some images are corrupted by substantial noise or artifacts.
In this work, we propose a novel approach based on a low-rank representation, i.e., Regnet-LRR, to tackle the problem.
We project noisy images into a noise-free low-rank space, and then compute the similarity between the images.
Based on the low-rank similarity measure, we train the registration network to predict the dense deformation fields of noisy image pairs.
We highlight that the low-rank projection is reformulated in a way that the registration network can successfully update gradients.
With two tasks, i.e., cardiac and abdominal intra-modality registration, we demonstrate that the low-rank representation can boost the generalization ability and robustness of models as well as bring significant improvements in noisy data registration scenarios.

\end{abstract}

\section{Introduction}\label{sec:introduction}

Pair-wise image registration aims at aligning one image to another.
Based on the registration, further medical image analysis can be conducted, including image-aided diagnosis, treatment and surgical decision-making.
In the last few decades, intensity-based registration methods have experienced extensive development and have been applied successfully to medical image analysis~\cite{ journal/tmi/Zhuang2011,Heinrich2012,journal/FN/Ashburner12,journal/tmi/Sotiras2013, journal/media/Viergever2016, journal/crp/Khalil2018, journal/media/Ashburner19}.
These traditional registration methods employ intensity-based similarity measures, including intensity difference (e.g., mean square error (MSE)) and correlation-based information (e.g., normalized cross correlation (NCC)), for better performance.

Deep neural networks (DNNs) are powerful in modeling complex functions and thus have been explored for image registration, which have achieved state-of-the-art accuracy~\cite{journal/media/Vos2018,journal/media/Hu2018,journal/tmi/Balakrishnan2019, journal/media/Dalca2019,journal/media/Sedghi20 }.
Some registration networks do not require ground truth of deformations or manual annotated segmentation, and Balakrishnan et al.~\cite{journal/tmi/Balakrishnan2019} named these methods unsupervised registration methods.
Unsupervised registration methods formulated the popular similarity measures as loss functions to optimize the network parameters~\cite{journal/media/Vos2018, journal/media/Hu2018, journal/media/Dalca2019}.

However, these similarity measures based on intensity differences can have a negative influence on the performance of unsupervised registration networks when substantial noise inevitably occurs in the image.
Since generic denoising models assume that noise is separable from the data, a simple idea to solve this problem is to perform registration after image denoising.
The two-stage strategy, including denoising and registration, usually requires two individual optimization processes to achieve the optimal solution and may be subject to computationally expensive iterations and parameter combinations.

Since data is inevitably accompanied with noise during the imaging process, denoising becomes essential in computer vision.
Owing to the rapid development of optimization techniques, many important denoising algorithms, such as model-based methods~\cite{Candes/2009,Candes_matrix/2010} and learning-based methods~\cite{Schmidt/2014,Dong/2016, Gao2020}, have been well explored.
Gu et al.~\cite{Gu/2014} employed low-rank components for image denoising by mining the self-similarity of images.

In this paper, we consider the scenario in which noisy images need to be registered. 
To the end, we propose an unsupervised image registration framework based on a low-rank representation, referred to as Regnet-LRR.
The model builds a novel loss function by computing the similarity between images in a low-rank space.
A convolutional neural network (CNN) is employed to achieve efficient optimization of the registration parameters.
Besides, Regnet-LRR performs registration of noisy images directly, thus avoiding performing registration after denoising.

The main contributions of this work are three-fold.
First, we resolve the problem of unsupervised registration for noisy images in an one-stage framework.
Second, we propose a novel loss function for registration which is derived from the low-rank theories. 
Third, by mapping the noisy images into low-rank space, our model outperforms previous intensity-based ones in pair-wise registration of noisy medical images.

\paragraph{\rm{\textbf{Related work.}}}
Intensity variations between images make registration difficult.
To address the problem, a general approach to registration is to employ similarity measures that are robust to those intensity variations.
These measures can be assumed as a functional or statistical intensity distribution, rather than an identical one.
Such robust results have been achieved by building on intensity differences and computing image gradient histograms with a wide range of features such as SIFT~\cite{Lowe2004Distinctive} and GLOH~\cite{Mikolajczyk2005A}.
However, these methods are sensitive to more intensity variations, such as multiple modalities or substantial noise.
To perform registration between multi-modal images, Wachinger et al.~\cite{Wachinger2011Entropy} mapped images to structural representations that were only dependent on the depicted structures instead of the intensities.
However, the design and construction of a noise-free representation for improving medical image registration networks in medical imaging has not been explored in depth.
\section{Methods}\label{sec:methods}
\begin{figure}[t]
  \centering
    \includegraphics[width=9cm, height=3.8cm]{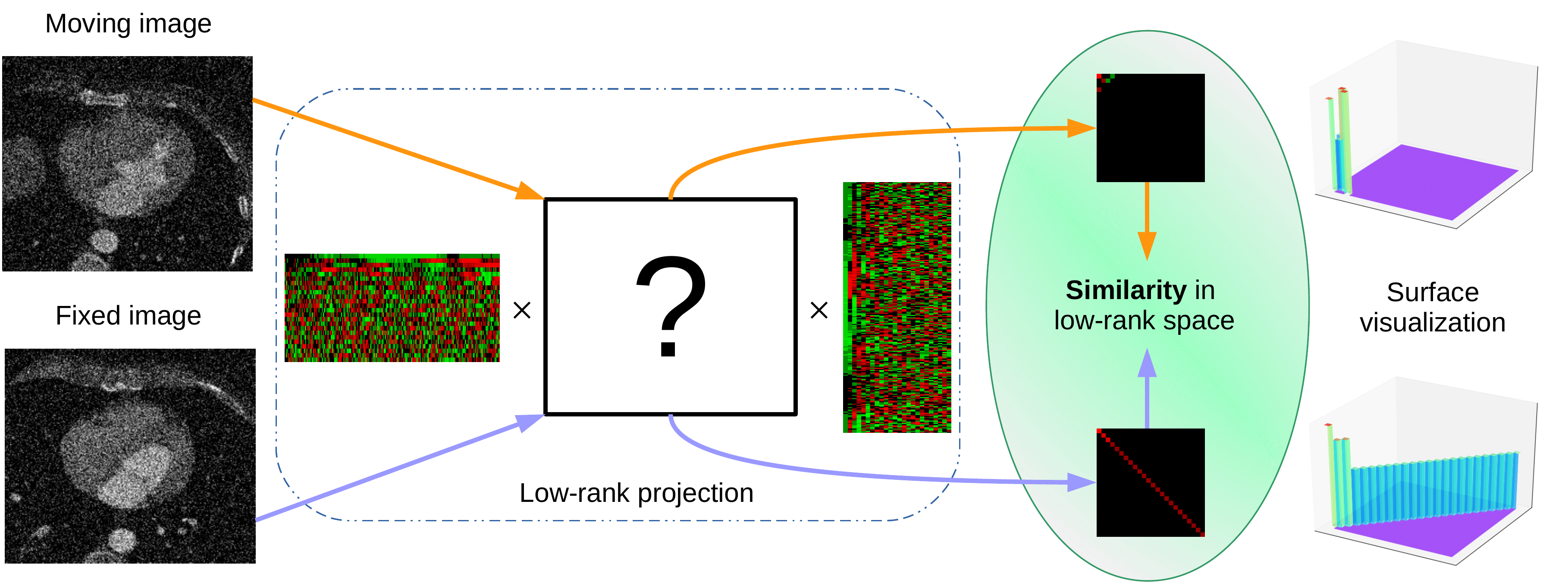}
  \caption{Schematic illustration of the registration framework with the low-rank representation. In this diagram, the procedure of low-rank projection of the noisy image pair is shown. Subsequently, the registration network is optimized with the similarity measure in the low-rank space.}
  \label{fig:ppl}
\end{figure}
A noisy intensity image can be considered as $\bm{I}: \Omega_{I} \rightarrow \mathbb{R}$, where $\Omega_{I} \subseteq \mathbb{R}^d$, and $d$ is the dimension of the image domain.
Registration of a noisy image pair ($\bm{I}_m$ and $\bm{I}_f$) can be modeled as a mapping from the fixed image domain $ \Omega_{\bm{I}_{f}}$ to the moving image domain $\Omega_{\bm{I}_m}$ by a transformation $\textbf{T}$.
Suppose $ \Theta $ is the parameter space for deformation mapping, the mapping $\textbf{T}$ could then be expressed as the elements in $ F = \left\lbrace \bm{T}_{\theta} | \bm{T}_{\theta} : \Omega_{\bm{I}_f} \rightarrow \Omega_{\bm{I}_m}, \theta\in \Theta\right\rbrace  $.
An optimal registration can be obtained via the optimization of the following objective function:
\begin{equation}\label{energy function}
\hat{\theta} \in \mathop{\arg\min}_{\theta \in \Theta}\mathcal{D}(\bm{I}_{m}\circ \textbf{T}_{\theta}, \bm{I}_{f}) + \lambda\cdot \mathcal{R}(\textbf{T}_\theta),
\end{equation}
where $\hat{\theta}$ is an element of the set of deformation parameters,
$\textbf{T}_{\theta}$ is the transformation mapping from the fixed image domain to the moving image domain.
In this paper, we assume the transformation $\textbf{T}_{\theta}$ as the voxel correspondence, which is represented by a dense deformation field (DDF).
$\mathcal{D}$ is a similarity measure between two image domains, and $\lambda\cdot\mathcal{R}(\textbf{T}_\theta)$ is the regularization term with weight $\lambda$.

\subsection{Image similarity measure in low-rank space}
\label{sec:ism}

Unsupervised registration is commonly performed based on the similarity measure $\mathcal{D}$ between two images.
However, since most of similarity measures are calculated directly in the original image space, they are quite sensitive to substantial noise in the images.

To tackle this problem, we propose the \textit{low-rank projection} method (\cref{sec:lrp}), which projects images into a noise-free low-rank space at first, and then computes the similarity in the space.

Since low-rank images have excellent self-similarity and noise-free property, we can calculate the image similarity after mapping the images to a low-rank space.
Suppose the low-rank component of an image can be obtained by the low-rank projection $ \bm{P}_r$, where $\mbox{rank}(\bm{P}_r(\bm{I})) \le r  $, we can then define the image similarity measure $\mathcal{D_{R}}$ between two images in the low-rank space as follows:
\begin{equation}\label{eq:low-rank measure}
\mathcal{D_{R}}(\bm{I}_{m}\circ \textbf{T}_{\theta}, \bm{I}_{f})=\|\bm{P}_r(\bm{I}_{m}\circ \textbf{T}_{\theta})-\bm{P}_r(\bm{I}_{f}) \|_F.
\end{equation}

We illustrate the process of computing the similarity measure in the low-rank space schematically in \cref{fig:ppl}.
Moreover, \cref{fig:lri} shows that using components with very low rank, i.e., $rank=5$, the consequent low-rank image is relatively similar to the clean image.
\subsection{Low-rank projection}
\label{sec:lrp}
\begin{figure}[t]
  \centering
  \includegraphics[width=0.85\textwidth]{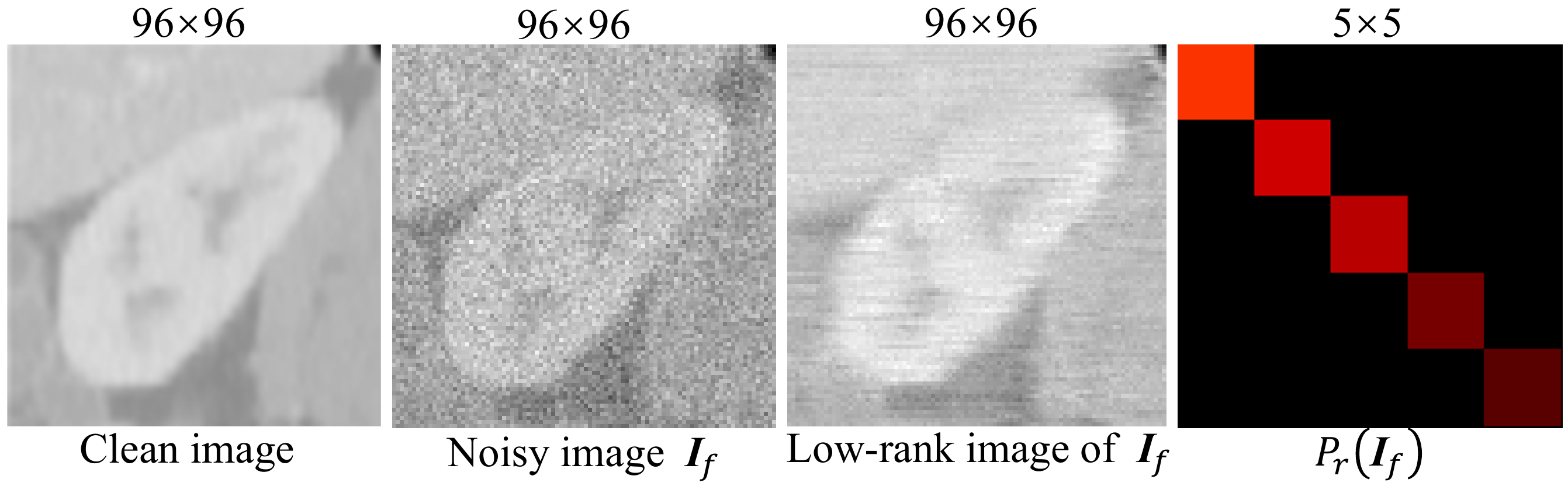}
  \caption{Visualization of the result of low-rank projection (rank=5) of a noisy image.
  This noisy image is used as the fixed image in Regnet-LRR.
  The low-rank projection of the noisy image $\bm{P}_r(\bm{I}_f)$ (\cref{sec:lrp}) is also visualized as an image.}
  \label{fig:lri}
\end{figure}
Singular value decomposition (SVD) can be a way to compute the low-rank components $ \bm{P}_r(\bm{I}_{m}\circ \textbf{T}_{\theta}) $ and $ \bm{P}_r(\bm{I}_{f}) $.
For example, suppose the SVD of $ \bm{I} $ is $ \bm{I} = U\Sigma V^\top $, then the low-rank component $ \bm{P}_r(\bm{I}) $ can be obtained by calculating $ U_r\Sigma_r V_r^\top $, where $ U_r $ is comprised of the first $ r $-th columns of $ U $, and $ V_r $ is comprised of the first $ r $-th columns of $ V $; $ \Sigma_r $ is a diagonal matrix whose diagonal elements are the first $ r $-th singular values. 

We could construct the low-rank projection $ \bm{P}_r(\bm{I})$ in two ways, i.e., SVD of the warped moving $ \bm{I}_{m}\circ \textbf{T}_{\theta} $ or SVD of the fixed image $ \bm{I}_{f}$.
But we found that directly computing the low-rank component of $ \bm{I}_{m}\circ \textbf{T}_{\theta} $ would make the network fail to back-propagate the gradients.

Therefore, we propose to build the low-rank projection $ \bm{P}_r $ by only using the SVD of $ \bm{I}_{f} $. 
Concretely, since $ U_r^\top \bm{I}_{f} V_r = \Sigma_r $, we can construct the low-rank projection in the following form:
\begin{equation}
\bm{P}_r(\bm{I}) = U_r^\top \bm{I} V_r.
\end{equation}
Here, $U_r$ and $V_r$ come from SVD of image $\bm{I}_f$.
Then, for any $ \bm{I} $, the rank of $ \bm{P}_r(\bm{I}) $ is not larger than $ r $, and can satisfy the constraint in the definition of $ \bm{P}_r $ (\cref{sec:ism}).
With such a low-rank projection, the formula (\ref{eq:low-rank measure}) can be converted to a new form as follow: 
\begin{equation}\label{eq:lowrankloss}
\mathcal{\hat{D}_{R}}(\bm{I}_{m}\circ \textbf{T}_{\theta}, \bm{I}_{f})=\|U_r^\top(\bm{I}_{m}\circ \textbf{T}_{\theta})V_r-U_r^\top(\bm{I}_{f})V_r \|_F=\|U_r^\top(\bm{I}_{m}\circ \textbf{T}_{\theta})V_r-\Sigma_r \|_F.
\end{equation}
We illustrate the low-rank projection schematically in \cref{fig:ppl}.
We also visualize the low-rank projection $ \bm{P}_r(\bm{I}_f) $ as a image in \cref{fig:lri}.

\subsection{Registration network with low-rank representation}
We employ a registration network which takes a noisy image pair as an input and estimates the DDF between the images, based on a 3D encoder-decoder-style architecture.
To encourage a suitable smooth displacement, we leverage bending energy as a deformation regularization term.
We combine the intensity-based measure with the weighted deformation regularization term as the loss function.
Hence, an optimal registration can be obtained via the optimization of the following objective function:
\begin{equation}\label{energy function}
\hat{\theta} \in \mathop{\arg\min}_{\theta \in \Theta}{\hat{\mathcal{D}}}_{R}(\bm{I}_{m}\circ \textbf{T}_{\theta}, \bm{I}_{f}) + \lambda\cdot \mathcal{R}(\textbf{T}_\theta).
\end{equation}

\section{Experiments and Results}\label{sec:experiments&results}

We demonstrated the registration effectiveness of our model by performing intra-modality pair-wise registration on two tasks, i.e., cardiac image registration and abdominal image registration.
Both tasks were performed on the synthetic noisy datasets.

\subsection{Material and experiment settings}
\subsubsection{Synthetic noisy datasets.}
The \textit{synthetic noisy datasets} were obtained by adding two types of noise commonly found in medical images, i.e., spatially invariant additive white Gaussian noise (AWGN) and Rician noise (RN), to \textit{clean}\footnote{Here, we assumed that the challenge data was noise-free for registration.
However, the cardiac and abdominal MRI images from challenges \cite{journal/media/Zhuang2019, 2020CHAOS} have degraded quality, due to artifacts and other factors.
Future work will study the issues.} intensity images, from publicly available medical image datasets.
Before adding noise, we employed intensity normalization to scale the intensity values to range [0,1].

Cardiac image registration was performed on the MM-WHS challenge dataset ~\cite{journal/media/Zhuang2016, journal/media/Zhuang2019}, which provides 40 labeled multi-modality whole-heart images from multiple sites, including 20 cardiac CT and 20 cardiac MRI.
We selected 10 subjects from each modality as training images. 

Abdominal image registration was performed on two multi-organ segmentation challenge datasets, i.e., Multi-Atlas Labeling Beyond the Cranial Vault challenge and CHAOS challenge~\cite{2020CHAOS}. 
The former consists of 30 labeled images scanned using CT modality, from which 20 images were chosen in random for training.
The latter consists of 20 labeled images scanned using MRI modality, from which 10 images were chosen in random for training.

\subsubsection{Evaluation.}
We evaluated the noisy image registration by calculating the Dice score between the label of warped moving image and the label of fixed image.
Cardiac image registration was evaluated by two substructures, i.e., myocardium (MYO) and left ventricle (LV), and abdominal image registration was evaluated by two organs i.e., right kidney (RK) and left kidney (LK).
We evaluated differences between Regnet-LRR and other methods on the same noisy datasets by Wilcoxon signed rank test \cite{Wilcoxon1944Individual} with a significance threshold of $p<0.05$.
\subsubsection{Experiment settings.}
In both tasks, we implemented and trained the registration networks via TensorFlow~\cite{arxiv/Abadi2015} on an $\text{NVIDIA}^\circledR$ $\text{RTX}^\textnormal{TM}$ 2080 Ti GPU.
The registration modules of Regnet-LRR were adapted from the open-source code including VoxelMorph~\cite{journal/tmi/Balakrishnan2019}, label-reg~\cite{journal/media/Hu2018}, MvMM-Regnet~\cite{Luo2020}, and DeepReg~\cite{Fu2020}.
Since Regnet-LRR estimated the 3D DDFs of images, we employed the Batch-SVD technique to optimize the network parameters.
The Batch-SVD module was adapted from the open-source code of ROnet~\cite{Gao2020}, and the pre-processing module was adapted from the open-source code of MvMM-Regnet~\cite{Luo2020} and CF-distance method~\cite{Wu2020CF}.
Moreover, the Adam optimizer~\cite{proceedings/ICLR/Kingma2014} was adopted, with a learning rate bouncing between 1e-5 and 1e-4 to accelerate convergence~\cite{proceedings/WACV/Smith2015}.

An optimal rank number results in a suitable low-rank projection, which minimizes the negative effect of noise on the registration results.
We set $r=48, \sigma=0.1$ as the default rank number and noise level, respectively.
Here, $\sigma$ denotes the standard deviation of AWGN and RN.
Moreover, high-quality registration depends on an optimal weight of bending energy which guarantees the global smoothness of the deformations.
To be balanced, we set $\lambda= 0.5$ as the default parameter of the regularization term.

We analyzed the registration effectiveness using different loss functions.
For example, we performed the registrations on noisy images using NCC loss function, referred to as G-Noisy-NCC (for AWGN noise), R-Noisy-NCC (for RN noise), and compared them with the same loss function on clean images, named as Clean-NCC, and so did MSE loss function.

\subsection{Task 1: intra-modality registration on noisy cardiac images}
\begin{table}[tbp]
  \small
  \centering
  \caption{Average substructure Dice (\%) of CT-to-CT and MRI-to-MRI pair-wise cardiac registration, with * indicating statistically significant improvement of Regnet-LRR on noisy datasets given by a Wilcoxon signed-rank test ($p<0.05$). Standard deviations are shown in parentheses.}
  \begin{tabular}{p{2cm}|L{1.6cm}|L{1.6cm}|L{1cm}|L{1.5cm}|L{1.5cm}|L{0.6cm}}
    \toprule
    \multirow{2}*{\textbf{Methods}} & \multicolumn{3}{C{4.2cm}|}{\textit{CT-to-CT}} & \multicolumn{3}{C{4.2cm}}{\textit{MRI-to-MRI}} \\
    \cline{2-7}
    & \multicolumn{1}{c|}{MYO} & \multicolumn{1}{c|}{LV}& \multicolumn{1}{c|}{Avg.} & \multicolumn{1}{c|}{MYO} & \multicolumn{1}{c|}{LV}& \multicolumn{1}{c}{Avg.} \\
    \hline
    Clean-MSE& $78.8(5.4)$ & $\textcolor{black}{83.7(5.1)}$ & $81.3$& $\textcolor{black}{70.2(5.2)}$ & $87.8(4.0)$& $79.0$ \\
    Clean-NCC& $75.6(11.1)$ & $82.8(9.5)$ &$79.2$& $\textcolor{black}{67.2(9.6)}$ & $88.5(3.7)$& $77.9$ \\
    \hdashline
    G-Noisy-MSE& $\textcolor{black}{73.6(5.2)}$ & $\textcolor{black}{77.8(4.8)}$&$75.7$ & $\textcolor{black}{65.1(5.5)}$ & $\textcolor{black}{86.0(4.7)}$& $75.6$ \\
    G-Noisy-NCC & $67.8(10.7)$ & $76.6(10.8)$ &$72.2$& $\textcolor{black}{65.3(10.8)}$ & $85.8(5.0)$& $75.6$ \\
    G-Noisy-LRR & $74.5(7.2)$* & $80.9(10.1)$*&$77.7$ & $67.8(5.2)$* & $87.1(3.3)$*& $77.5$ \\
    \hdashline
    R-Noisy-MSE& $\textcolor{black}{73.9(5.4)}$ & $\textcolor{black}{77.3(6.5)}$&$75.6$ & $\textcolor{black}{66.1(6.5)}$ & $\textcolor{black}{86.6(5.9)}$& $76.4$ \\
    R-Noisy-NCC & $68.8(11.1)$ & $77.0(10.0)$ &$72.9$& $\textcolor{black}{66.2(10.6)}$ & $83.0(6.9)$& $74.6$ \\
    R-Noisy-LRR & $75.6(7.4)$* & $82.2(10.5)$*&$78.9$ & $68.5(4.9)$* & $87.2(3.6)$*& $77.9$ \\
    \bottomrule
  \end{tabular}
  \label{tab:heart_reg_dice}
\end{table}

Registration was performed after noise was added to the clean data.
This setting simulated a practical noisy image registration problem, in which registration of noisy images was more difficult than that of clean data.

As shown in \cref{tab:heart_reg_dice}, while the registration achieved excellent performance on clean data (Clean-MSE and Clean-NCC), its performance dropped in noisy cases.
For example, due to negative effect of the AWGN noise, the average Dice of myocardium and left ventricle dropped from 79.2\% to 72.2\% using NCC loss function with respect to CT modality.
Obviously, intensity-based registration methods had problems due to the negative effect of certain level of noise.

By contrast, with the noise-free low-rank representation, we observed better registration accuracy on noisy CT and MRI images.
Moreover, our method outperformed both the MSE and NCC loss functions on noisy images, yielding substantial and consistent improvements on two cardiac substructures.
Notably, Regnet-LRR attained evident improvements on challenging cardiac MRI images with Rician noise, where the shape of the myocardium was clearly irregular, reaching an average myocardium Dice of 68.5\% compared with 66.1\% (R-Noisy-MSE) and 66.2\% (R-Noisy-NCC).

\subsection{Task 2: intra-modality registration on noisy abdominal images}

In this experiment, we evaluated the proposed method with respect to the abdominal datasets with both CT and MRI modalities.
Since the left and right kidneys are separate organs on the abdominal image, we cropped them out for registration separately.

\begin{table}[tbp]
  \small
  \centering
  \caption{Average organ Dice (\%) of CT-to-CT and MRI-to-MRI pair-wise abdomen registration, with * indicating statistically significant improvement of Regnet-LRR on noisy dataset given by a Wilcoxon signed-rank test ($p<0.05$). Standard deviations are shown in parentheses.}
    \begin{tabular}{p{2cm}|L{1.5cm}|L{1.5cm}|L{1cm}|L{1.5cm}|L{1.5cm}|L{0.6cm}}
    \toprule
    \multirow{2}*{\textbf{Methods}} & \multicolumn{3}{C{4cm}|}{\textit{CT-to-CT}} & \multicolumn{3}{C{4cm}}{\textit{MRI-to-MRI}} \\
    \cline{2-7}
    & \multicolumn{1}{c|}{RK} & \multicolumn{1}{c|}{LK}& \multicolumn{1}{c|}{Avg.} & \multicolumn{1}{c|}{RK} & \multicolumn{1}{c|}{LK}& \multicolumn{1}{c}{Avg.} \\
    \hline
    Clean-MSE& $82.6(4.8)$ & $84.6(3.9)$ & $83.6$& $82.7(5.3)$ & $80.0(8.7)$& $81.4$ \\
    Clean-NCC& $79.1(3.1)$ & $86.9(2.9)$ &$83.0$& $84.8(3.6)$ & $83.1(7.6)$& $84.0$ \\
    \hdashline
    G-Noisy-MSE& $78.9(5.1)$ & $81.5(3.8)$&$80.2$ & $\textcolor{black}{80.6(4.9)}$ & $\textcolor{black}{76.9(8.6)}$& $78.8$ \\
    G-Noisy-NCC & $69.4(6.6)$ & $70.5(5.7)$ &$70.0$& $77.7(6.6)$ & $\textcolor{black}{78.2(5.7)}$& $78.0$ \\
    G-Noisy-LRR & $82.1(4.4)$* & $83.2(7.9)$*&$82.7$ & $83.8(4.0)$* & $80.1(8.8)$*& $82.0$ \\
    \hdashline
    R-Noisy-MSE& $78.5(7.1)$ & $81.7(6.2)$&$80.1$ & $\textcolor{black}{80.6(5.9)}$ & $\textcolor{black}{78.1(9.1)}$& $79.4$ \\
    R-Noisy-NCC & $70.9(6.8)$ & $69.4(5.2)$ &$70.2$& $79.0(5.0)$ & $80.0(5.5)$& $79.5$ \\
    R-Noisy-LRR & $82.1(4.6)$* & $84.3(3.5)$*&$83.2$ & $83.3(4.4)$* & $81.8(8.7)$*& $82.6$ \\
    \bottomrule
  \end{tabular}
  \label{tab:reg_dice_abdomen}
\end{table}

\cref{tab:reg_dice_abdomen} compares the registration accuracy obtained by different loss functions on clean and noisy images.
The intensity-based loss functions were less competitive on the noisy datasets than the clean datasets, e.g., the average Dice of G-Noisy-NCC on abdominal images in CT modality dropped from 83.0\% to 70.0\%.
On the other hand, our framework achieved higher performance in pair-wise abdominal image registration on noisy datasets.

\subsection{Ablation study}

\begin{figure}[t]
  \centering
  \includegraphics[width=8cm, height=3.5cm]{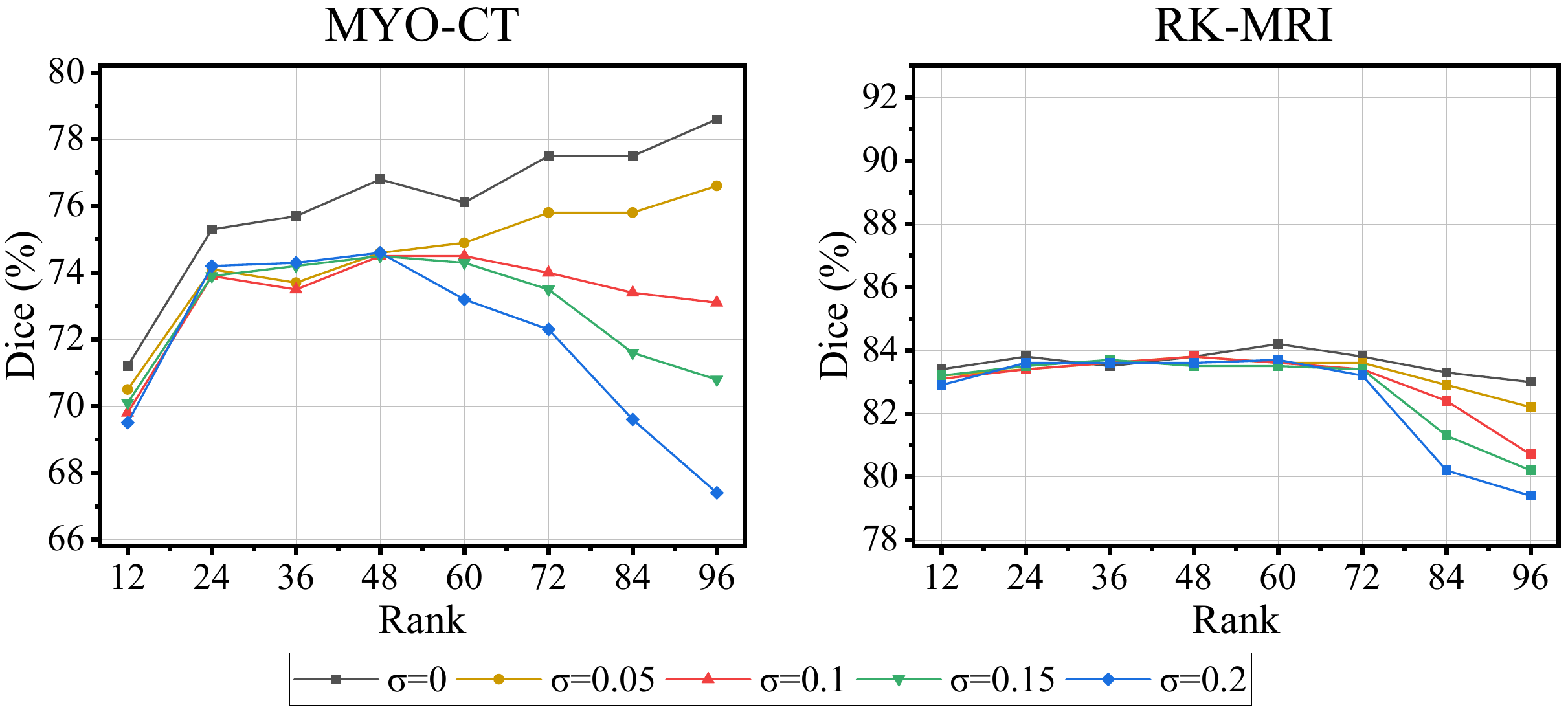}
  \caption{Average substructure/organ Dice scores of pairwise labels using Regnet-LRR with $\sigma$ noise levels and different rank numbers.}
  \label{fig:reg_dice_abdomen}
\end{figure}

To better understand the effectiveness of the low-rank representation for noisy image registration, we compared it with different noise levels and rank numbers.
We added AWGN with specific noise levels, i.e., $\sigma \in [0,0.2]$, and then made the rank number vary from 12 to 96.
\cref{fig:reg_dice_abdomen} reported the mean Dice scores, i.e., myocardium Dice with respect to CT modality (MYO-CT) and Dice of right kidney with respect to MRI modality (RK-MRI), obtained from Regnet-LRR.

The results clearly showed that the low-rank representation improved the robustness to noise with respect to intensity-based unsupervised registration models.
When having a rank smaller than a specific threshold, i.e., $rank\le48$ for MYO-CT, and $rank\le72$ for RK-MRI, Regnet-LRR was insensitive to different noise levels.
These results aimed to demonstrate that the low-rank representation can keep the image registration free from certain level of noise due to the advantages of image self-similarity.

However, we also observed growing sensitivity of the low-rank representation to the noise level as the rank increases.
This discrepancy could be attributed to the incorporation of more high frequency information from the image as more intensity noises were also encompassed.

\section{Conclusion}
In this work, we propose a low-rank representation and formulate a novel loss function for the unsupervised registration network on noisy medical images.
We have evaluated the proposed model on two tasks, i.e. cardiac and abdominal pair-wise registration based on three synthetic noisy datasets generated from publicly available medical image datasets.
The proposed approach has shown its noise-free efficacy in registration on noisy medical images.

\bibliographystyle{splncs04}
\bibliography{paper}

\end{document}